\documentclass[final]{cvpr}

\usepackage{times}
\usepackage{epsfig}
\usepackage{graphicx}
\usepackage{subfigure}
\usepackage{caption}
\usepackage{amsmath}
\usepackage{amssymb}
\usepackage{multirow}
\usepackage{color}
\usepackage{booktabs}
\usepackage{bbding}
\usepackage{makecell}

\usepackage[pagebackref,breaklinks,colorlinks]{hyperref}

\usepackage[capitalize]{cleveref}
\crefname{section}{Sec.}{Secs.}
\Crefname{section}{Section}{Sections}
\Crefname{table}{Table}{Tables}
\crefname{table}{Tab.}{Tabs.}

\graphicspath{{./figs/}{./figs/result/}}

\begin{document}

\title{PP-YOLOE-R: An Efficient Anchor-Free Rotated Object Detector}

\author{Xinxin Wang, Guanzhong Wang, Qingqing Dang, Yi Liu, Xiaoguang Hu, Dianhai Yu \\
Baidu Inc.\\
\tt\small \{wangxinxin08, wangguanzhong, dangqingqing, liuyi22, huxiaoguang, yudianhai\} @baidu.com
}

\maketitle

\begin{abstract}

  Arbitrary-oriented object detection is a fundamental task in visual scenes involving aerial images and scene text. In this report, we present PP-YOLOE-R, an efficient anchor-free rotated object detector based on PP-YOLOE. We introduce a bag of useful tricks in PP-YOLOE-R to improve detection precision with marginal extra parameters and computational cost. As a result, PP-YOLOE-R-l and PP-YOLOE-R-x achieve 78.14 and 78.28 mAP respectively on DOTA 1.0 dataset with single-scale training and testing, which outperform almost all other rotated object detectors. With multi-scale training and testing, PP-YOLOE-R-l and PP-YOLOE-R-x further improve the detection precision to 80.02 and 80.73 mAP. In this case, PP-YOLOE-R-x surpasses all anchor-free methods and demonstrates competitive performance to state-of-the-art anchor-based two-stage models. Further, PP-YOLOE-R is deployment friendly and PP-YOLOE-R-s/m/l/x can reach 69.8/55.1/48.3/37.1 FPS respectively on RTX 2080 Ti with TensorRT and FP16-precision. Source code and pre-trained models are available at PaddleDetection$\footnote{\scriptsize\url{https://github.com/PaddlePaddle/PaddleDetection}\label{ppdet}}$, which is powered by
  PaddlePaddle$\footnote{\scriptsize\url{https://github.com/PaddlePaddle/Paddle}\label{paddle}}$.
  
\end{abstract}

\section{Introduction}

\begin{figure}[ht]
\centering
\includegraphics[width=\linewidth]{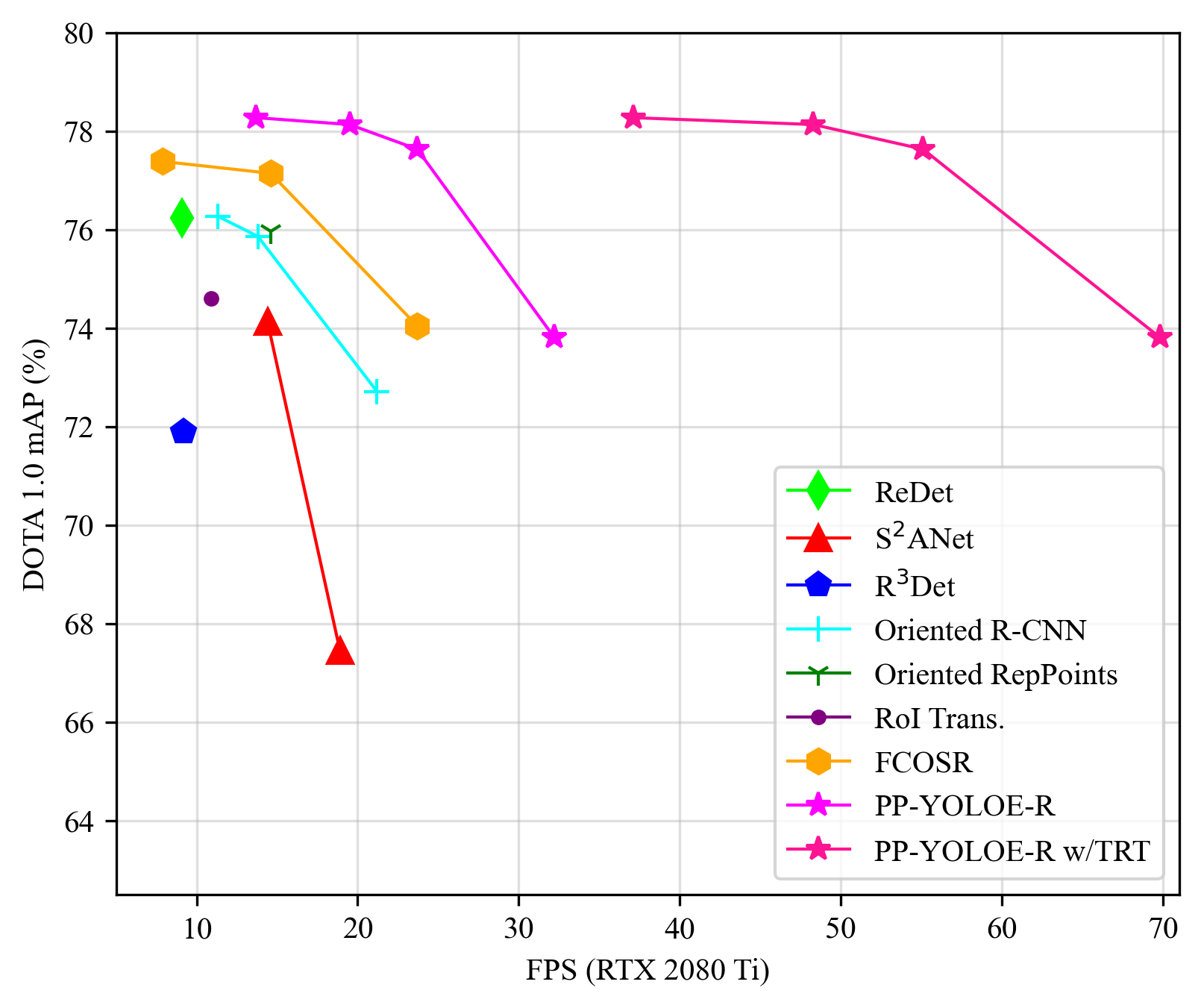} 
\caption{Comparison of PP-YOLOE-R and other state-of-the-art models with single-scale training and testing on DOTA 1.0 dataset. PP-YOLOE-R-s/m/l/x achieves 73.82/77.64/78.14/78.28 mAP respectively at the speed of 69.8/55.1/48.3/37.1 FPS on RTX 2080 Ti with TensorRT and FP16-precision.}
\label{figure1}
\end{figure}

\begin{figure*}[ht]
	\centering
	\includegraphics[width=\textwidth]{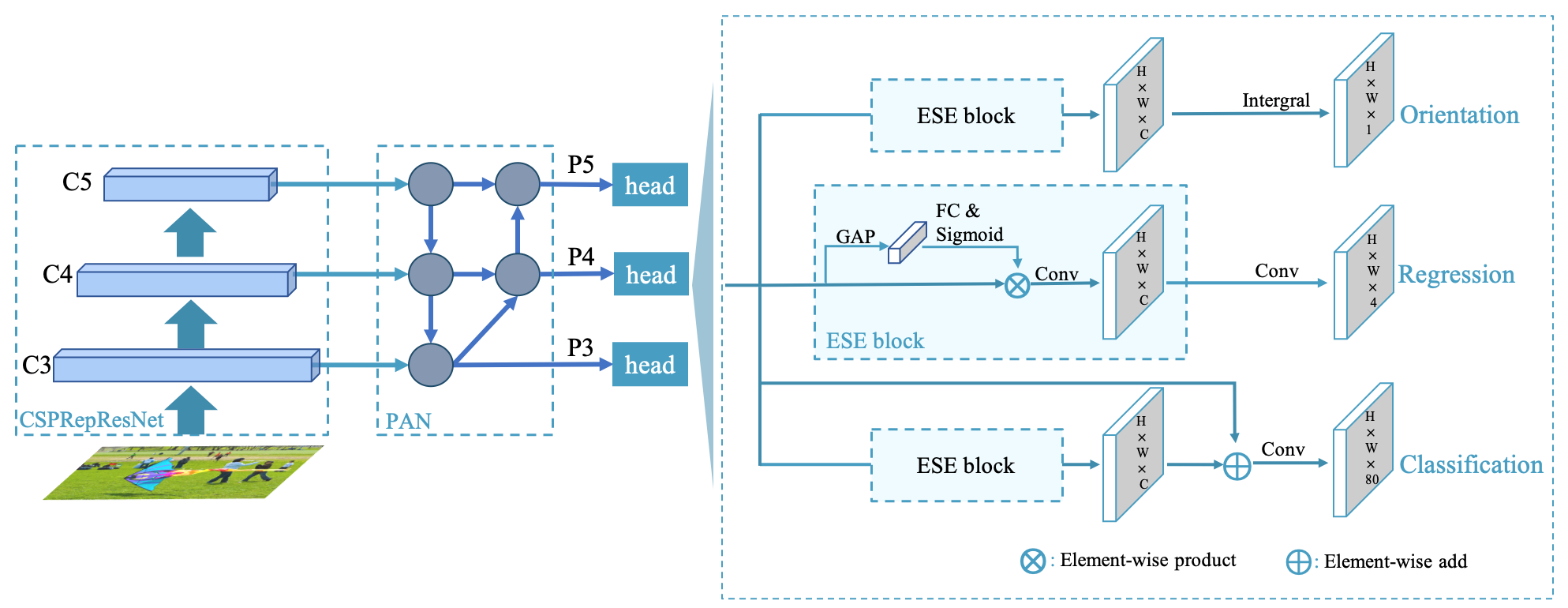}
	\caption{The overall architecture of PP-YOLOE-R. The structure of PP-YOLOE-R is similar to that of PP-YOLOE, except that a decoupled angle prediction head is introduced into PP-YOLOE-R.}
	\label{figure2}
\end{figure*}

Detecting arbitrary-oriented objects is significant to understand remote sensing images and has attracted increasing attention. Due to massive variations in the scale and orientation of objects, rotated object detection still remains challenging. Benefiting from the rapid development of the horizontal object detection, more and more rotated object detectors\cite{ding2019learning, guo2021beyond,  han2021redet, yang2021r3det, han2021align, ming2021dynamic, li2021fcosr, lang2021dafne, hou2022shape, li2022oriented, xie2021oriented} have emerged gradually, which are mainly derived from corresponding horizontal object detectors\cite{lin2017focal, ren2015faster, tian2019fcos, yang2019reppoints}. Among these rotated object detectors, the representation of oriented objects can be roughly divide into three ways, which are rotated bounding boxes with five parameters, quadrangles with eight parameters and a set of key points. Currently, rotated object detectors based on five-parameter representation dominate this research area. Although having achieved promising result, direct five-parameter regression still has some theoretical problem such as boundary discontinuity problem\cite{yang2021dense, yang2020arbitrary, yang2021rethinking}.
The boundary discontinuity problem is mainly caused by periodicity of angle and exchange ability of edges while the latter is relative to specific definition of rotated bounding boxes such as long-edge definition. There are a lot of works proposed to resolve the boundary discontinuity problem such as \cite{llerena2021gaussian, yang2021rethinking, yang2021learning, yang2022kfiou, yang2021dense, yang2020arbitrary}. \cite{llerena2021gaussian, yang2021rethinking, yang2021learning, yang2022kfiou} model rotated bounding box as Gaussian distribution and propose computing-friendly IoU-based loss as a substitute for differentiable SkewIoU loss to avoid direct angle regression. \cite{yang2021dense, yang2020arbitrary} consider angular prediction as classification and design smooth label to avoid boundary discontinuity problem. Fully draw on the excellent ideas of advanced horizontal and oriented detectors, we propose PP-YOLOE-R, an efficient anchor-free rotated object detector based on PP-YOLOE\cite{xu2022pp}

Compared with PP-YOLOE, the main changes of PP-YOLOE-R can be attributed to four aspects: (1) we introduce ProbIoU loss\cite{llerena2021gaussian} like \cite{li2021fcosr} as regression loss to avoid boundary discontinuity problem. (2) we introduce Rotated Task Alignment Learning to be suitable for rotated object detection on the basis of Task Alignment Learning\cite{feng2021tood} (3) we design a decoupled angle prediction head and directly learn the General distribution of angle through DFL loss\cite{li2020generalized} for a more accurate angle prediction. (4) we make a slight modification to the re-parameterization mechanism\cite{ding2021repvgg} by adding a learnable gating unit to control the amount of information from the previous layer. As a result, PP-YOLOE-R achieves state-of-the-art performance in terms of speed and accuracy trade-off on DOTA 1.0 dataset. Specifically, PP-YOLOE-R-l and PP-YOLOE-R-x achieve 78.14 and 78.28 mAP respectively with single-scale training and testing. With multi-scale training and testing, PP-YOLOE-R-l and PP-YOLOE-R-x further improve the detection precision to 80.02 and 80.73 mAP respectively. While maintaining high precision, PP-YOLOE-R-l can achieve the speed of 48.3 FPS at 1024$\times$1024 resolution with TensorRT and FP16-precision. Moreover, PP-YOLOE-R-s and PP-YOLOE-R-m also have excellent performance and are suitable for edge devices with relatively low computing power. Our code can be available at PaddleDetection\cite{ppdet2019}. 

\section{Related work}

{\bf Anchor-Based Rotated Object Detectors.} Anchor-based rotated object detectors have one-stage and two-stage approches similar to horizontal object detectors. RoI Transformer\cite{ding2019learning} proposes a RRoI learner to predict the offsets of Rotated Ground Truths (RGTs) relative to predicted HRoI. Oriented R-CNN\cite{xie2021oriented} designs a lightweight oriented RPN to generate high-quality oriented proposals. ReDet\cite{han2021redet} introduces rotation-equivariant CNN (ReCNN) to obtain rotation-equivariant feature maps and Rotation-invariant RoI Align (RiRoI Align) to extract features of RRoIs. S$^2$ANet\cite{han2021align} and R$^3$Det\cite{yang2021r3det} both adopt the refined single-stage framework to detect oriented objects. S$^2$ANet proposes Alignment Convolution Layer (ACL) based on Deformable Convolution Network while R$^3$Det designs Feature Refinement Module (FRM) based on interpolation to alleviate feature misalignment.

{\bf Anchor-Free Rotated Object Detectors.} Anchor-free rotated object detectors are mainly based on center point or a set of key points. DAFNe\cite{lang2021dafne} proposes oriented centerness and center-to-corner prediction strategy while FCOSR\cite{li2021fcosr} focuses on the label assignment strategy based on FCOS\cite{tian2019fcos} to improve detection performance. CFA\cite{guo2021beyond} and Oriented RepPoints\cite{li2022oriented} indirectly predict oriented bounding boxes by predicting nine representative points based on RepPoints\cite{yang2019reppoints}.

\begin{table*}[ht]
	\begin{center}
		\begin{tabular}{l|l|c|c}
			\hline
			Model & mAP(\%) & Parameters(M) & GFLOPs \\
			\hline \hline
			
			baseline & 75.61 & 50.65 & 269.09  \\
			\hline
			
			 +Rotated Task Alignment Learning & 77.24 (\textcolor[RGB]{225,10,10}{\small\textbf{$+$1.63}}) &    50.65  & 269.09     \\
			 +Decoupled Angle Prediction Head & 77.78 (\textcolor[RGB]{225,10,10}{\small\textbf{$+$0.54}}) &    52.20  & 272.72        \\			
			 +Angle Prediction with DFL       & 78.01 (\textcolor[RGB]{225,10,10}{\small\textbf{$+$0.23}}) &    53.29  & 281.65        \\
			 +Learnable Gating Unit for RepVGG   &  \textbf{78.14} (\textcolor[RGB]{225,10,10}{\small\textbf{$+$0.13}}) & 53.29 & 281.65  \\
			\hline
		\end{tabular}
	\end{center}
	
	\caption{Ablation study of PP-YOLOE-R-l on DOTA 1.0 dataset with single-scale training and testing. All parameters and GFLOPs are calculated after re-parameterization and the resolution of the input image is 1024$\times$1024}
	\label{roadmap}
\end{table*}

{\bf Label Assignment.} The purpose of label assignment is to distinguish positive and negative samples. Label assignment can be divide into static and dynamic strategies. During training, dynamic label assignment utilizes the output of model as a basis for selecting positive and negative samples while static label assignment determines positive and negative samples according to ground truths and predefined rules. FCOSR\cite{li2021fcosr} proposes ellipse center sampling method, fuzzy sample assignment strategy and multi-level sampling module to ease insufficient sampling prolem. \cite{hou2022shape} proposes Shape-Adaptive Selection (SA-S) to adjust IoU threshold according to the shape of samples. G-Rep\cite{hou2022g} substitutes IoU with normalized Gaussian distribution distance as an assignment indicator. DAL\cite{ming2021dynamic} introduces a matching degree considering a prior of spatial matching and feature alignment ability to dynamic select positive samples. Similarly, Oriented RepPoints\cite{li2022oriented} designs a quality measure to assign samples.

{\bf Loss.} Due to periodicity of angle and exchange ability of edges, direct regression-based rotated object detectors suffer boundary discontinuity problem. CSL\cite{yang2020arbitrary} and DCL\cite{yang2021dense} predict angle in a classification way. To avoid boundary discontinuity problem, GWD\cite{yang2021rethinking}, ProbIoU\cite{llerena2021gaussian}, KLD\cite{yang2021learning} and KFIoU\cite{yang2022kfiou} converts the rotated bounding box to a 2D Gaussian distribution and constructs a distance metric of two Gaussian distributions to measure the similarity of two rotated bounding boxes. GWD\cite{yang2021rethinking} utilizes Gaussian Wasserstein distance to  approximate SkewIoU while ProbIoU\cite{llerena2021gaussian} utilizes Bhattacharyya Coefficient to measure the similarity of two rotated bounding boxes. KLD\cite{yang2021learning} calculates Kullback-Leibler Divergence (KLD) between two Gaussian distributions as the regression loss. Moreover, KFIoU\cite{yang2022kfiou} achieves trend-level alignment with SkewIoU by adopting Kalman filter to mimic SkewIoU according to its definition. 

\section{Method}
As shown in \cref{figure2}, the overall architecture of PP-YOLOE-R is similar to that of PP-YOLOE. PP-YOLOE-R improves detection performance of rotated bounding boxes at the expense of relatively small amount of parameters and computation based on PP-YOLOE. In this section, we will introduce the changes made for rotated bounding boxes in detail.

\noindent {\bf Baseline.} Drawing on FCOSR\cite{li2021fcosr}, we introduce FCOSR Assigner and ProbIoU Loss into PP-YOLOE as our baseline. FCOSR Assigner is adopted to assign ground truth to three feature maps according to the predefined rules and ProbIoU Loss is utilized as regression loss. The backbone and neck of our baseline remain the same as PP-YOLOE while the regression branch of head is modified to predict five-parameter rotated bounding boxes ($x, y, w, h, \theta$) directly. Our baseline achieves 75.61 mAP with single-scale training and testing as shown in \Cref{roadmap}.

\noindent {\bf Rotated Task Alignment Learning.} Task Alignment Learning\cite{feng2021tood} is composed of a task-aligned label assignment and task-aligned loss. Task-aligned label assignment constructs a task alignment metric to select positive samples from candidate anchor points, whose coordinates fall into any ground truth boxes. The task alignment metric is calculated as follows:
\begin{equation}
    t = s^\alpha \cdot \mu^\beta
\end{equation}
where $s$ denotes a predicted classification score and $\mu$ denotes the IoU value between predicted bounding box and corresponding ground truth. In Rotated Task Alignment Learning, the selection process of candidate anchor points takes advantage of the geometric properties of the ground truth bounding box and anchor points in it and the SkewIoU value of predicted and ground truth bounding box is adopted as $\mu$. With these two simple changes, we can apply task-aligned label assignment for rotated object detection and use task-aligned loss without modification. By using Rotated Task Alignment Learning, the detection precision is further improved to 77.24 mAP as shown in \Cref{roadmap}. 

\noindent {\bf Decoupled Angle Prediction Head.} In the regression branch, most rotated object detectors predict five parameters ($x, y, w, h, \theta$) to represent the oriented object. However, we assume that predicting $\theta$ requires different features than predicting ($x, y, w, h$). To verify this hypothesis, we design a decoupled angle prediction head as shown in \Cref{figure2} to predict $\theta$ and ($x, y, w, h$) separately. The angle prediction head consists of a channel attention layer and a convolution layer and is very lightweight. By introducing the decoupled angle prediction head, the detection precision is improved by 0.54 mAP to 77.24 mAP as shown in \Cref{roadmap}.

\begin{table*}[ht]
    \centering
    \resizebox{1.0\textwidth}{!}{
        \begin{tabular}{c|c|c|ccccccccccccccc|c}
        \hline \hline
        \textbf{Methods} & \textbf{Backbone} & \textbf{MS} & \textbf{PL} & \textbf{BD} & \textbf{BR} & \textbf{GTF} & \textbf{SV} & \textbf{LV} & \textbf{SH} & \textbf{TC} & \textbf{BC} & \textbf{ST} & \textbf{SBF} & \textbf{RA} & \textbf{HA} & \textbf{SP} & \textbf{HC} & \textbf{mAP}\\
        \hline
        \textit{\textbf{Anchor-based Methods}} & & &
        & & & & & & & & & &
        & & & & &\\
        RoI-Trans.\cite{ding2019learning} & R101 & \checkmark & 
        88.64 & 78.52 & 43.44 & 75.92 & 68.81 & 73.68 & 83.59 & 90.74 & 77.27 & 81.46 & 
        58.39 & 53.54 & 62.83 & 58.93 & 47.67 & 69.56 \\
        DAL\cite{ming2021dynamic} & R101 &  &
        88.61 & 79.69 & 46.27 & 70.37 & 65.89 & 76.10 & 78.53 & 90.84 & 79.98 & 78.41 &
        58.71 & 62.02 & 69.23 & 71.32 & 60.65 & 71.78 \\
        CSL\cite{yang2020arbitrary} & R152 & \checkmark &
        \textcolor[RGB]{225,10,10}{\textbf{90.25}} & \textcolor[RGB]{225,10,10}{\textbf{85.53}} & 54.64 & 75.31 & 70.44 & 73.51 & 77.62 & 90.84 & 86.15 & 86.69 &
        69.60 & 68.04 & 73.83 & 71.10 & 68.93 & 76.17 \\
        R$^3$Det\cite{yang2021r3det} & R152 & \checkmark &
        89.80 & 83.77 & 48.11 & 66.77 & 78.76 & 83.27 & 87.84 & 90.82 & 85.38 & 85.51 &
        65.57 & 62.68 & 67.53 & 78.56 & 72.62 & 76.47 \\
        DCL\cite{yang2021dense} & R152 & \checkmark &
        89.26 & 83.60 & 53.54 & 72.76 & 79.04 & 82.56 & 87.31 & 90.67 & 86.59 & 86.98 &
        67.49 & 66.88 & 73.29 & 70.56 & 69.99 & 77.37 \\
        S$^2$ANet\cite{han2021align} & R50 & \checkmark &
        88.89 & 83.60 & 57.74 & 81.95 & 79.94 & 83.19 & \textcolor[RGB]{225,10,10}{\textbf{89.11}} & 90.78 & 84.87 & 87.81 &
        70.30 & 68.25 & 78.30 & 77.01 & 69.58 & 79.42 \\
        ReDet\cite{han2021redet} & ReR50 & \checkmark & 
        88.81 & 82.48 & 60.83 & 80.82 & 78.34 & 86.06 & 88.31 & 90.87 & \textcolor[RGB]{225,10,10}{\textbf{88.77}} & 87.03 &
        68.65 & 66.90 & 79.26 & \textcolor[RGB]{225,10,10}{\textbf{79.71}} & 74.67 & 80.10 \\
        GWD\cite{yang2021rethinking} & R152 & \checkmark &
        89.66 & 84.99 & 59.26 & 82.19 & 78.97 & 84.83 & 87.70 & 90.21 & 86.54 & 86.85 & 
        \textcolor[RGB]{225,10,10}{\textbf{73.47}} & 67.77 & 76.92 & 79.22 & 74.92 & 80.23 \\
        KLD\cite{yang2021learning} & R152 & \checkmark & 
        89.92 & 85.13 & 59.19 & 81.33 & 78.82 & 84.38 & 87.50 & 89.80 & 87.33 & 87.00 & 
        72.57 & 71.35 & 77.12 & 79.34 & \textcolor[RGB]{225,10,10}{\textbf{78.68}} & 80.63 \\
        Oriented R-CNN\cite{xie2021oriented} & R50 & \checkmark & 
        89.84 & 85.43 & 61.09 & 79.82 & 79.71 & 85.35 & 88.82 & 90.88 & 86.68 & 87.73 & 
        72.21 & 70.80 & \textcolor[RGB]{225,10,10}{\textbf{82.42}} & 78.18 & 74.11 & 80.87 \\
        RoI-Trans. + KFIoU\cite{yang2022kfiou} & Swin-Tiny & \checkmark &
        89.44 & 84.41 & \textcolor[RGB]{225,10,10}{\textbf{62.22}} & \textcolor[RGB]{225,10,10}{\textbf{82.51}} & \textcolor[RGB]{225,10,10}{\textbf{80.10}} & \textcolor[RGB]{225,10,10}{\textbf{86.07}} & 88.68 & \textcolor[RGB]{225,10,10}{\textbf{90.90}} & 87.32 & \textcolor[RGB]{225,10,10}{\textbf{88.38}} &
        72.80 & \textcolor[RGB]{225,10,10}{\textbf{71.95}} & 78.96 & 74.95 & 75.27 & \textcolor[RGB]{225,10,10}{\textbf{80.93}} \\
        \hline
        \textit{\textbf{Anchor-free Methods}} & & &
        & & & & & & & & & &
        & & & & & \\
        BBAVectors\cite{yi2021oriented} & R101 & \checkmark &
        88.63 & 84.06 & 52.13 & 69.56 & 78.26 & 80.40 & 88.06 & 90.87 & 87.23 & 86.39 &
        56.11 & 65.62 & 67.10 & 72.08 & 63.96 & 75.36  \\
        CFA\cite{guo2021beyond} & R152 & \checkmark &
        89.08 & 83.20 & 54.37 & 66.87 & 81.23 & 80.96 & 87.17 & 90.21 & 84.32 & 86.09 &
        52.34 & 69.94 & 75.52 & 80.76 & 67.96 & 76.67 \\
        Oriented RepPoints\cite{li2022oriented} & Swin-Tiny &  &
        89.11 & 82.32 & 56.71 & 74.95 & 80.70 & 83.73 & 87.67 & 90.81 & 87.11 & 85.85 &
        63.60 & 68.60 & 75.95 & 73.54 & 63.76 & 77.63 \\
        FCOSR-s\cite{li2021fcosr} & MV2 & &
        89.09 & 80.58 & 44.04 & 73.33 & 79.07 & 76.54 & 87.28 & \textcolor[RGB]{225,10,10}{\textbf{90.88}} & 84.89 & 85.37 &
        55.95 & 64.56 & 66.92 & 76.96 & 55.32 & 74.05 \\
        FCOSR-s\cite{li2021fcosr} & MV2 & \checkmark &
        88.60 & 84.13 & 46.85 & 78.22 & 79.51 & 77.00 & 87.74 & 90.85 & 86.84 & 86.71 &
        64.51 & 68.17 & 67.87 & 72.08 & 62.52 & 76.11 \\
        FCOSR-m\cite{li2021fcosr} & X50 & &
        88.88 & 82.68 & 50.10 & 71.34 & 81.09 & 77.40 & 88.32 & 90.80 & 86.03 & 85.23 &
        61.32 & 68.07 & 75.19 & 80.37 & 70.48 & 77.15 \\
        FCOSR-m\cite{li2021fcosr} & X50 & \checkmark &
        89.06 & 84.93 & 52.81 & 76.32 & 81.54 & 81.81 & 88.27 & 90.86 & 85.20 & 87.58 &
        68.63 & \textcolor[RGB]{225,10,10}{\textbf{70.38}} & 75.95 & 79.73 & 75.67 & 79.25 \\
        FCOSR-l\cite{li2021fcosr} & X101 & &
        \textcolor[RGB]{225,10,10}{\textbf{89.50}} & 84.42 & 52.58 & 71.81 & 80.49 & 77.72 & 88.23 & 90.84 & 84.23 & 86.48 &
        61.21 & 67.77 & 76.34 & 74.39 & 74.86 & 77.39 \\
        FCOSR-l\cite{li2021fcosr} & X101 & \checkmark &
        88.78 & \textcolor[RGB]{225,10,10}{\textbf{85.38}} & 54.29 & 76.81 & 81.52 & 82.76 & 88.38 & 90.80 & 86.61 & 87.25 &
        67.58 & 67.03 & 76.86 & 73.22 & 74.68 & 78.80 \\
        \textbf{PP-YOLOE-R-s} & CRN-s & &
        88.80 & 79.24 & 45.92 & 66.88 & 80.41 & 82.95 & 88.20 & 90.61 & 82.91 & 86.37 &
        55.80 & 64.11 & 65.09 & 79.50 & 50.43 & 73.82 \\
        \textbf{PP-YOLOE-R-s} & CRN-s & \checkmark &
        88.93 & 83.95 & 56.60 & \textcolor[RGB]{225,10,10}{\textbf{79.40}} & 82.57 & 85.89 & 88.64 & 90.87 & 87.82 & 87.54 & 
        68.94 & 63.46 & 76.66 & 79.19 & 70.87 & 79.42 \\
        \textbf{PP-YOLOE-R-m} & CRN-m & &
        89.23 & 79.92 & 51.14 & 72.94 & 81.86 & 84.56 & 88.68 & 90.85 & 86.85 & 87.48 & 
        59.16 & 68.34 & 73.78 & 81.72 & 68.10 & 77.64 \\
        \textbf{PP-YOLOE-R-m} & CRN-m & \checkmark &
        88.63 & 84.45 & 56.27 & 79.12 & \textcolor[RGB]{225,10,10}{\textbf{83.52}} & \textcolor[RGB]{225,10,10}{\textbf{86.16}} & 88.77 & 90.81 & 88.01 & \textcolor[RGB]{225,10,10}{\textbf{88.39}} & 
        \textcolor[RGB]{225,10,10}{\textbf{70.41}} & 61.44 & 77.65 & 77.70 & 74.30 & 79.71 \\
        \textbf{PP-YOLOE-R-l} & CRN-l & &
        89.18 & 81.00 & 54.01 & 70.22 & 81.85 & 85.16 & 88.81 & 90.81 & 86.99 & 88.01 & 
        62.87 & 67.87 & 76.56 & 79.13 & 69.65 & 78.14 \\
        \textbf{PP-YOLOE-R-l} & CRN-l & \checkmark &
        88.40 & 84.75 & 58.91 & 76.35 & 83.13 & 86.10 & 88.79 & 90.87 & \textcolor[RGB]{225,10,10}{\textbf{88.74}} & 87.71 & 
        67.71 & 68.44 & \textcolor[RGB]{225,10,10}{\textbf{77.92}} & 76.17 & 76.35 & 80.02 \\
        \textbf{PP-YOLOE-R-x} & CRN-x & &
        89.49 & 79.70 & 55.04 & 75.59 & 82.40 & 85.20 & 88.35 & 90.76 & 85.69 & 87.70 &
        63.17 & 69.52 & 77.09 & 75.08 & 69.38 & 78.28 \\
        \textbf{PP-YOLOE-R-x} & CRN-x & \checkmark &
        88.45 & 84.46 & \textcolor[RGB]{225,10,10}{\textbf{60.57}} & 77.70 & 83.34 & 85.36 & \textcolor[RGB]{225,10,10}{\textbf{88.97}} & 90.78 & 88.53 & 87.47 & 
        69.26 & 65.96 & 77.86 & \textcolor[RGB]{225,10,10}{\textbf{81.36}} & \textcolor[RGB]{225,10,10}{\textbf{80.93}} & \textcolor[RGB]{225,10,10}{\textbf{80.73}} \\
        \hline
        \end{tabular}
    }
	\vspace{0.1cm}
	\caption{Comparison with state-of-the-art methods on DOTA 1.0 dataset. R50 and X50 denote ResNet-50 and ResNeXt-50 (likewise for R101, R152 and X101). MV2 denotes MobileNetv2 and CRN denotes CSPRepResNet. MS means multi-scale training and testing. The bold \textcolor[RGB]{225,10,10}{\textbf{red}} fonts indicate the best performance.}
	\label{tab-state-of-the-art}
\end{table*}

\noindent {\bf Angle Prediction with DFL.} ProbIoU Loss is adopted as regression loss to jointly optimize ($x, y, w, h, \theta$). To calculate ProbIoU Loss, the rotated bounding box is converted to a Gaussian bounding box. When the rotated bounding box is roughly square, the orientation of the rotated bounding box cannot be determined because the orientation in Gaussian bounding box is inherited from the elliptical representation. To overcome this problem, we introduce Distribution Focal Loss (DFL) \cite{li2020generalized} to predict the angle. Different form $l_n$-norm learning the Dirac delta distribution, DFL is aimed at learning the General distribution of angle. Specifically, we discretize the angle with even intervals $\omega$ and obtain the preidcted $\theta$ in the form of integral, which can be formulated as follows:
\begin{equation}
    \theta = \sum_{i=0}^{90} p_i \cdot i \cdot \omega
\end{equation}
where $p_i$ means the probability that the angle falls in each interval. In this paper, the rotated bounding box is defined in OpenCV definition and $\omega$ is set to $\pi/180$. By introducing DFL, the detection precision is improved by 0.23 mAP to 78.01 mAP.

\noindent {\bf Learnable Gating Unit for RepVGG.} RepVGG proposes a multi-branch architecture composed of a 3$\times$3 conv, a 1$\times$1 conv and a shortcut path. The training-time information flow of RepVGG can be formulated as follows:
\begin{equation}
    y = f(x) + g(x) + x
\end{equation}
while $f(x)$ is a 3$\times$3 conv and $g(x)$ is a 1$\times$1 conv. During inference, we can re-parameterizes this architecture to an equivalent 3$\times$3 conv. Although RepVGG is equivalent to a convolution layer, using RepVGG during training converges better. We attribute this result to that the design of RepVGG introduces useful prior knowledge. Inspired by this, we introduce a learnable gating unit in RepVGG to control the amount of information from previous layer. This design is mainly for tiny objects or dense objects to adaptively fuse the features with different receptive fields, which can be formulated as follows:
\begin{equation}
    y = f(x) + \alpha_1 \cdot g(x) + \alpha_2 \cdot x
\end{equation}
where $\alpha_1$ and $\alpha_2$ are learnable parameters. In our RepResBlock\cite{xu2022pp}, the shortcut path is not used so we introduce only one parameter for each RepResBlock. During inference, the learnable parameter can be re-parameterized along with convolution layers so that neither the speed nor the amount of parameters changes. By introducing learnable gating unit, the detection precision is improved by 0.13 mAP to 78.14 mAP.

\noindent {\bf ProbIoU Loss.} By modeling rotated bounding boxes as Gaussian bounding boxes, the Bhattacharyya Coefficient of two Gaussian distributions is used to measure the similarity of two rotated bounding boxes in ProbIoU\cite{llerena2021gaussian}. GWD\cite{yang2021rethinking}, KLD\cite{yang2021learning} and KFIoU \cite{yang2022kfiou} are also similarity measures based on Gaussian bounding boxes. To verify the effect of ProbIoU loss, we choose KLD loss for experiment because KLD loss is scale invariant and suitable for anchor-free methods. As shown in \Cref{loss_ablation}, replacing ProbIoU loss with KLD loss causes a significant preformance drop from 78.14 mAP to 76.03 mAP, which indicates that ProbIoU loss is more suitable for our design.

\begin{table}[ht]
	\begin{center}
		\begin{tabular}{c|c}
			\hline
			Loss & mAP(\%) \\
			\hline
			
			ProbIoU loss\cite{llerena2021gaussian} & \textbf{78.14}  \\
			KLD loss\cite{yang2021learning} & 76.03 \\
			\hline
		\end{tabular}
	\end{center}
	
	\caption{Ablation study of different loss on PP-YOLOE-R-l}
	\label{loss_ablation}
\end{table}

\section{Experiment}

\subsection{Dataset}
DOTA\cite{xia2018dota} is a large-scale remote sensing dataset for oriented object detection, which contains 15 categories: plane (PL), baseball diamond (BD), bridge (BR), ground track field (GTF), small vehicle (SV), large vehicle (LV), ship (SH), tennis court (TC), basketball court (BC), storage tank (ST), soccer ball field (SBF), roundabout (RA), harbor (HA), swimming pool (SP), and helicopter (HC). DOTA is comprised of 2806 aerial images with the size about 4000 × 4000 pixels and 188,282 instances with a wide variety of scales, orientations, and shapes. Half of the aerial images are randomly selected as the training set, 1/6 as the validation set, and 1/3 as the testing set. For single-scale training and testing, we crop the original images into 1024$\times$1024 patches with an overlap of 256 pixels. For multi-scale training and testing, the original images are resized with the scale of 0.5, 1.0 and 1.5 and then cropped into 1024$\times$1024 patches with an overlap of 500 pixels.
\subsection{Implementation details}
PP-YOLOE-R adopts CSPRepResNet as backbone and PAN as neck to extract P3, P4 and P5 pyramid features for rotated object detection. The stochastic gradient descent (SGD) with momentum = 0.9 and weight decay = 5e-4 is used in our training. The initial learning rate is set to 0.008 with the warming up for 1000 iterations and cosine learning rate schedule is used after warming up. We train all the models with 36 epochs for DOTA 1.0 dataset and use 4 Tesla V100 GPU devices with 32G memory for training with the total batch size of 8. During training process, the exponential moving average (EMA) strategy with decay = 0.9998 is also adopted. We adopt random flip and adopt a two step rotation augmentation method following FCOSR\cite{li2021fcosr} to generate random augmentation data.

\subsection{Comparsion with Other SOTA Detectors}
We conduct extensive experiments on the DOTA 1.0 dataset and the experimental results are shown in \Cref{tab-state-of-the-art}. With single-scale training and testing, PP-YOLOE-R-l and PP-YOLOE-R-x achieves 78.14 and 78.28 mAP respectively, which outperform almost all rotated object detectors. With multi-scale training and testing, PP-YOLOE-R-l and PP-YOLOE-R-x further improve the detection precision to 80.02 and 80.73 mAP. PP-YOLOE-R-x outperforms all anchor-free methods and is only 0.2 mAP lower than the two-stage anchor-based model with the highest precision. Moreover, PP-YOLOE-R-s and PP-YOLOE-R-m can achieve 79.42 and 79.71 mAP with multi-scale training and testing, which are excellent results considering the parameters and GLOPS of these two models. While maintaining high precision, PP-YOLOE-R avoids using special operators, such as Deformable Convolution or Rotated RoI Align, to be deployed friendly on various hardware. As a result, PP-YOLOE-R can be easily accelerated with TensorRT whereas most other SOTA models are currently not easy to deploy using TensorRT. At the input resolution of 1024$\times$1024, PP-YOLOE-R-s/m/l/x can reach 32.2/23.7/19.5/13.7 FPS on RTX 2080Ti. With TensorRT and FP-16 precision, PP-YOLOE-R-s/m/l/x can be further accelerated to 69.8/55.1/48.3/37.1 FPS respectively.

\section{Conclusion}
In this report, we propose PP-YOLOE-R, an efficient anchor-free rotated object detector based on PP-YOLOE. PP-YOLOE-R achieves high precision and real-time speed with marginal extra parameters and computational cost, surpassing all anchor-free rotated object detectors. PP-YOLOE-R is easy to deploy and has a series of models for different computing power devices, named s/m/l/x. In the future, we will conduct experiments on more datasets for rotated object detection and extend PP-YOLOE-R to related scenes.

{\small
\bibliographystyle{ieee_fullname}
\bibliography{egbib}
}

\end{document}